\documentclass[conference]{IEEEtran}
\IEEEoverridecommandlockouts
\usepackage{cite}
\usepackage{amsmath,amssymb,amsfonts}
\usepackage{algorithmic}
\usepackage{graphicx}
\usepackage{textcomp}
\usepackage{xcolor}
\def\BibTeX{{\rm B\kern-.05em{\sc i\kern-.025em b}\kern-.08em
    T\kern-.1667em\lower.7ex\hbox{E}\kern-.125emX}}

\makeatletter
\newcommand{\linebreakand}{%
  \end{@IEEEauthorhalign}
  \hfill\mbox{}\par
  \mbox{}\hfill\begin{@IEEEauthorhalign}
}
\makeatother

\usepackage[caption=false,font=small]{subfig}

\DeclareMathOperator{\argmin}{argmin}

\usepackage{diagbox}
\usepackage{multirow}
\usepackage{color, colortbl}
\definecolor{Gray}{gray}{0.9}

\usepackage{array}
\newcolumntype{P}[1]{>{\centering\arraybackslash}p{#1}}
\newcolumntype{M}[1]{>{\centering\arraybackslash}m{#1}}

\begin{document}

\title{Learning-enabled multi-modal motion prediction in urban environments\\
\thanks{This work has been partially funded by the Spanish Ministry of Science and  Innovation with the National Project NEWCONTROL (PCI2019-103791), the Community of Madrid through SEGVAUTO 4.0-CM Programme (S2018-EMT-4362), and by the European Commission and ECSEL Joint Undertaking through the Project NEWCONTROL (826653).}
}






\author{\IEEEauthorblockN{Vinicius Trentin}
\IEEEauthorblockA{\textit{Center for Automation and Robotics} \\
\textit{CSIC - Universidad Politécnica de Madrid} \\
Arganda del Rey, Spain \\
vinicius.trentin@csic.es}
\and
\IEEEauthorblockN{Chenxu Ma}
\IEEEauthorblockA{\textit{Computer Engineering Lab } \\
\textit{TU Delft } \\
Delft, The Netherlands \\
c.ma-6@student.tudelft.nl}
\and
\IEEEauthorblockN{Jorge Villagra}
\IEEEauthorblockA{\textit{Center for Automation and Robotics} \\
\textit{CSIC - Universidad Politécnica de Madrid} \\
Arganda del Rey, Spain \\
jorge.villagra@csic.es}
\linebreakand
\IEEEauthorblockN{Zaid Al-Ars}
\IEEEauthorblockA{\textit{Computer Engineering Lab} \\
\textit{TU Delft} \\
Delft, The Netherlands \\
Z.Al-Ars@tudelft.nl}
}

\maketitle

\begin{abstract}

Motion prediction is a key factor towards the full deployment of autonomous vehicles. It is fundamental in order to assure safety while navigating through highly interactive complex scenarios. In this work, the framework IAMP (Interaction-Aware Motion Prediction), producing multi-modal probabilistic outputs from the integration of a Dynamic Bayesian Network and Markov Chains, is extended with a learning-based approach. The integration of a machine learning model tackles the limitations of the ruled-based mechanism since it can better adapt to different driving styles and driving situations. The method here introduced generates context-dependent acceleration distributions used in a Markov-chain-based motion prediction. This hybrid approach results in better evaluation metrics when compared with the baseline in the four highly-interactive scenarios obtained from publicly available datasets.

\end{abstract}


\section{Introduction}


 The anticipation of possible dangerous driving situations is fundamental to take preventive actions and to minimize potential risks accordingly. Indeed, in order to perform safe and efficient motion planning, autonomous vehicles need to predict the evolution of other traffic participants. However, the behavior of the surrounding agents is full of uncertainties in the real world and depends on the layout of the driving scene and on their interactions with others \cite{villagra2023motion}. As a result, a reliable and robust mechanism for intention estimation and motion prediction is critical for autonomous vehicles.

The motion prediction problem has a multi-modal nature since a vehicle has many possible trajectories while navigating through the layout. To solve the multi-modal motion prediction problem, two categories of solutions are usually applied: model-based and learning-based. Model-based methods try to mathematically formalize the dynamics of the problem, whereas in learning-based approaches, the interactions and dynamics of the system are learned from data.

Model-based approaches include stochastic filters. In \cite{Yonghwan}, the author predicts the probabilistic future states of the surrounding vehicles and the uncertainty of these predictions via a self-adaptive motion predictor. He does so by considering a kinematic model of the vehicles' motion and a Kalman filter to estimate the uncertainty. \cite{Lefkopoulos} proposed a motion prediction scheme based on an Interactive Multiple Model Kalman Filter that is able to infer the intentions and generate interaction-aware non-colliding predictions of multiple vehicles considering a priority list. Due to the uncertainty produced by the Kalman Filter, these methods should only be used to predict short prediction horizons. To analyze the interaction between vehicles and predict their route and maneuver intentions, the authors in \cite{Schulz} employed a Dynamic Bayesian Network with a particle filter. An action, indicated by acceleration and yaw rate values, is derived from these intentions, and the motion prediction is computed. This technique evaluates just the most likely action for the whole prediction horizon, which may have a detrimental impact on the motion planning search space in complicated scenarios.

In previous works from the authors \cite{vinicius_access, Vinicius_Springer}, they introduced a framework based on the combination of a Dynamic Bayesian Network to infer intentions and a Markov chain-based motion prediction to estimate the motion of the surrounding vehicles. Although the multi-modal model-based framework previously implemented can be successfully applied to any driving situation or scenario, it cannot adapt to different driving styles. 

In recent years, many neural network architectures have been employed to model the surrounding context and interactions among traffic agents in motion prediction. \cite{Gu} use a sparse context encoder to extract agent and map attributes, a dense goal encoder to retrieve the goal probability distribution, and a goal set predictor to provide predicted trajectories based on the goal probability distribution. \cite{THOMAS} describes a cooperative multi-agent trajectory prediction system that uses a graph encoder and a grid decoder to make goal-based predictions. It takes the agents' history and a lanelet map as inputs and generates a heatmap of potential ending locations, from which \textit{K} trajectories are created by decoding \textit{K} end points and generating the path with a fully connected neural network. \cite{Yicheng} presents a network architecture based on a stacked transformer for modeling multi-modal predictions using predetermined trajectories collected from traffic data as input. \cite{Gao} predicts the route of the target vehicles using a hierarchical graph neural network, which receives vector representations of agents and maps as input and retrieves interactions between agents via a fully-connected graph. \cite{Chai} employs trajectory anchors from the training dataset to model intent uncertainty using a Gaussian Mixture Model whose parameters are generated using a deep neural network. Notice that despite the very promising prediction results of most of these methods, data-driven approaches rely on a vast amount of data, and the results obtained with them can degrade when applied to different scenarios from those selected for their training. Furthermore, these methods are usually based on complex machine learning systems that have limited interpretability.

In this paper, the previous work \cite{vinicius_access, Vinicius_Springer} used to infer intentions and compute the motion prediction of the surrounding vehicles is extended to consider a learning-enabled motion prediction. The neural network model implemented provides personalized acceleration profiles from the most relevant information about the vehicle and its interactions with the layout and with other vehicles. These profiles serve as input to the Markov chain-based motion prediction, resulting in a hybrid approach that combines both model- and learning-based techniques. It can be applied to urban scenarios with high interactions where the autonomous vehicle needs accurate predictions in order to safely navigate, such as roundabouts and intersections. The extended framework is compared with the baseline in four situations obtained from publicly available datasets.

The outline of this paper is organized as follows: Section \ref{sec: iamp_framework} presents a review of IAMP framework. Section~\ref{sec: learning_models} introduces the model implemented to generate the acceleration profiles. Section~\ref{sec: experimental_results} presents the scenarios being used and shows the results for the 4 driving environments. Finally, Section~\ref{sec: conclusion} provides some concluding remarks.

\section{Interaction-aware motion prediction}
\label{sec: iamp_framework}


The interaction-aware motion prediction uses a combination of Dynamic Bayesian Network (DBN) and Markov chains to infer the intentions and predict the motion of the surrounding vehicles. It explicitly considers interactions, road layout, and traffic rules.

At each time step, the vehicle-to-vehicle and vehicle-to-layout interactions are taken into account to infer the probability to stop or to cross the intersections, the probability to change lanes, as well as the probability of being in each of the possible navigable corridors. These intentions are fused with the motion predictions computed with a kinematic model to result in a motion grid used by the ego vehicle to navigate through the scene.

The grid-based representation of the predictions takes into account the uncertainties both in the motion model and in the input data, resulting in a more reliable and robust prediction when compared with a point-based trajectory prediction.

\begin{figure}[!t]
    \centering
    \includegraphics[width=0.8\linewidth]{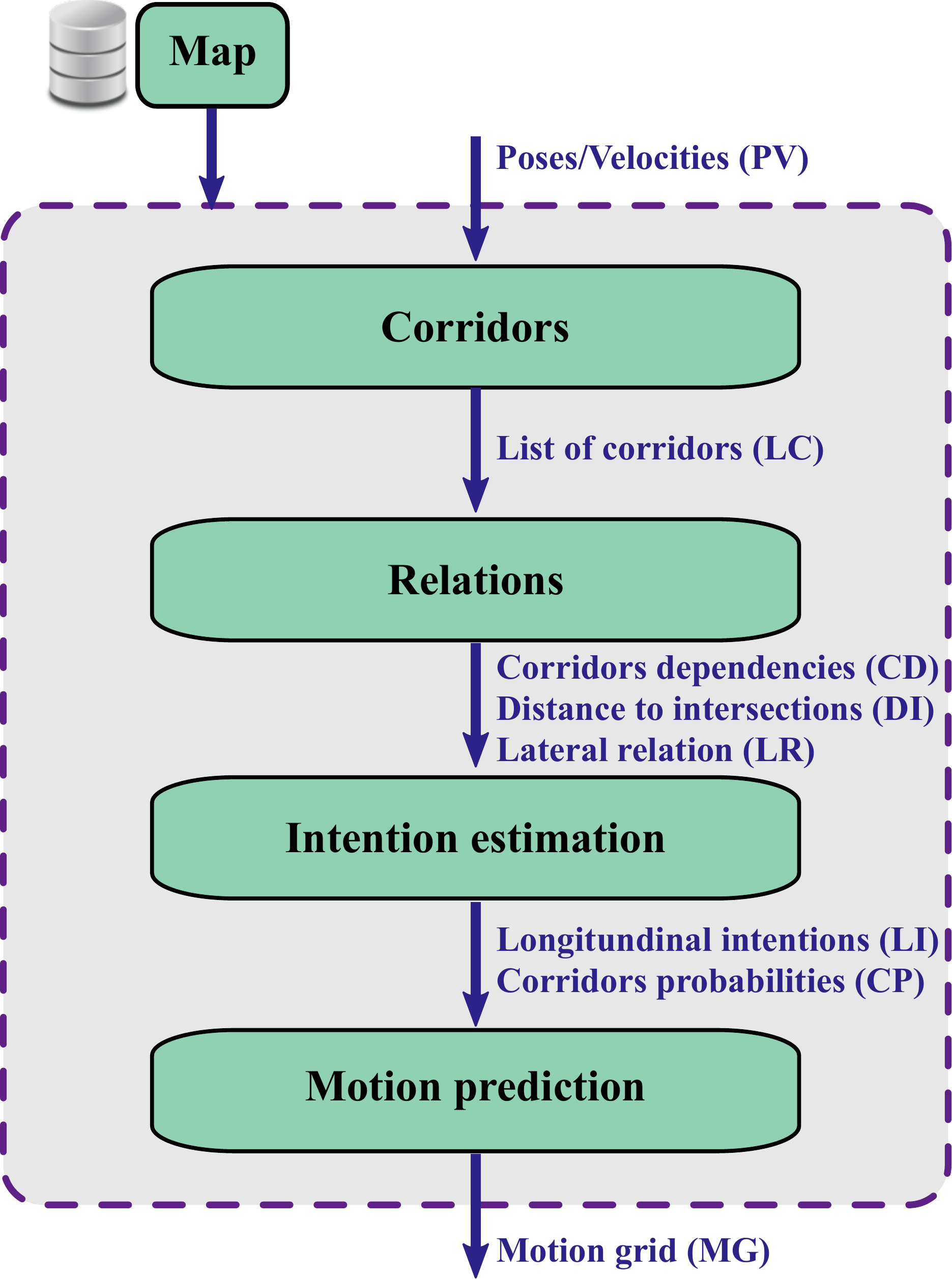}
    \caption{Interaction-aware motion prediction flowchart.}
    \label{fig: flowchart}
\end{figure}

The framework can be divided in four main blocks: \textit{Corridors}, \textit{Relations}, \textit{Intentions} and \textit{Motion Prediction}. An overview of the framework is presented in Figure \ref{fig: flowchart} where the data entering and leaving each block is shown. Each block will be briefly described below. For more details, the reader is referred to \cite{vinicius_access, Vinicius_Springer}.

\subsection{Maps}

The maps are loaded at the beginning of the simulation. They are made up of lanelets \cite{lanelets}, which are interconnected driveable segments geometrically represented by a left and a right bound. The regulatory elements, which are linked to the lanelets, can indicate which lanelets have right of way, which must yield, and the exact placement of the stop lines in the case of an unsignalized intersection.

\subsection{Corridors}

A corridor is a lanelet sequence that represents a possible route a vehicle might take. They are obtained from the physical and relational layers extracted from the lanelets. The extension of these corridors is limited to the distance that the vehicle can travel in a given time interval at its current velocity, assuming constant acceleration. 

\subsection{Relations}

The interactions obtained from the list of corridors and the map are threefold: lateral relation, corridor-to-intersection (distance to intersections) and corridor-to-corridor (corridors dependencies);

\begin{itemize}

    \item lateral relation: for each target vehicle, a search of the surrounding vehicles is performed, and the bumper-to-bumper distance and velocities are stored.
    
    \item corridor-to-intersection: the intersections each corridor goes through are determined by intersecting the lanelet's identifiers. For all intersections a corridor goes through, the distance to the intersection is computed in the Frenet frame. The entrance through which the corridor passes is also determined.
    
    \item corridor-to-corridor: determine which corridor will influence the predictions of the other acting as an obstacle ahead. The centerlines of all corridors are pairwise intersected, generating a list of possible collisions. Based on a set of rules, one (if any) corridor is selected as the corridor influencing the motion of a vehicle in a given corridor.
    
\end{itemize}

\subsection{Intentions}

\begin{figure}[!t]
    \centering
    \includegraphics[width=0.45\linewidth]{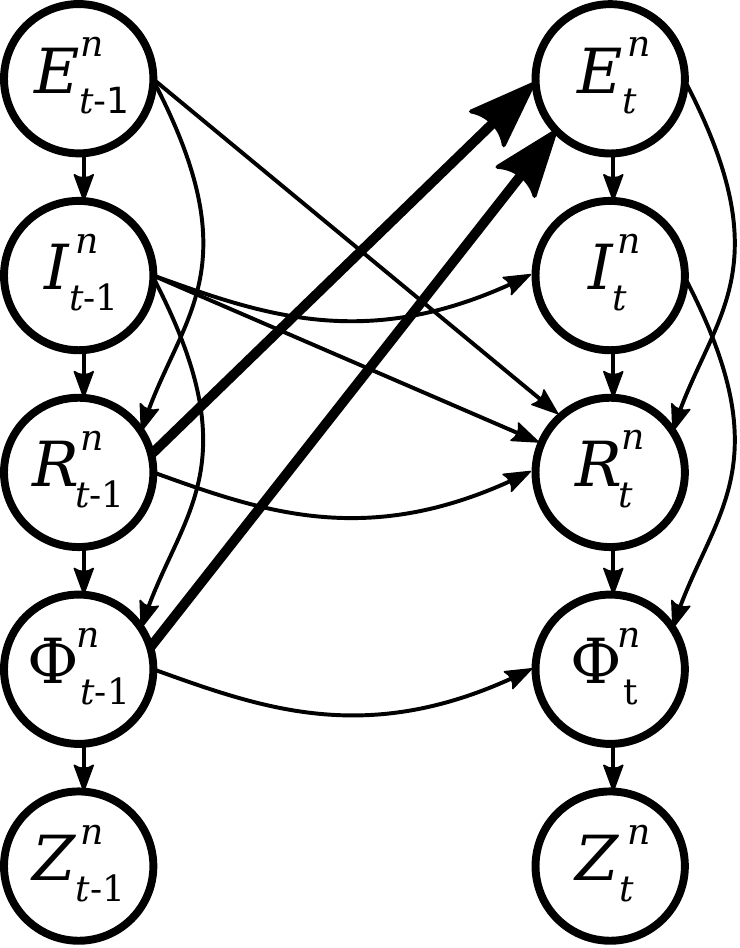}
    \caption{Bayesian network for two consecutive time steps.}
    \label{fig:bayesianNet}
\end{figure}

The Dynamic Bayesian Network (DBN) described in \cite{lefevre} and implemented in \cite{vinicius_access,Vinicius_Springer} is used to estimate the intention of traffic participants. The network represented in Figure \ref{fig:bayesianNet} is instantiated for all vehicles present in the scene, where bold arrows represent the influences of the other vehicles on vehicle $n$ through some key variables  ($E_{t}^{n}$, $I_{t}^{n}$, $R_{t}^{n}$, $\Phi_{t}^{n}$, $Z_{t}^{n}$).

\begin{itemize}
    \item Expected maneuver $E_{t}^{n}$: reflects the expected behavior of the vehicle $n$ at moment $t$ in accordance with traffic laws.
    
    \item Intended maneuver $I_{t}^{n}$: reflects the intention of the vehicle.

    \item Route $R_{t}^{n}$: estimated route the vehicle is following.
    
    \item Physical vehicle state $\Phi_{t}^{n}$: represents the pose, curvature, and speed of the vehicle. They are determined at each instant based on the vehicles' intentions.
    
    \item Measurements $Z_{t}^{n}$: represents the real measurements of the physical state of the vehicle, obtained directly from exteroceptive sensors of the EV or via V2X communications \cite{godoy2021}. 

\end{itemize}

The relationships between the variables in Fig. \ref{fig:bayesianNet} allow the following generalized distribution to be used to model the driving scene:

\begin{equation}
    \begin{gathered}
    P(\boldsymbol{E_{0:T}}, \boldsymbol{I_{0:T}}, \boldsymbol{R_{0:T}}, \boldsymbol{\Phi_{0:T}}, \boldsymbol{Z_{0:T}}) = P(\boldsymbol{E_{0}}, \boldsymbol{I_{0}}, \boldsymbol{R_{0}}, \boldsymbol{\Phi_{0}}, \boldsymbol{Z_{0}}) \times \\ \prod_{t=1}^{T} \times  \prod_{n=1}^{N} [P(E_{t}^{n}|\boldsymbol{R_{t-1},\Phi_{t-1}}) \times P(I_{t}^{n}|I_{t-1}^{n},E_{t}^{n}) \times \\ P(\Phi_{t}^{n}|\Phi_{t-1}^{n},R_{t}^{n},I_{t}^{n}) \times P(Z_{t}^{n}|\Phi_{t}^{n})]
    \label{eq: bayesian_distribution}
    \end{gathered}
\end{equation}

Since an exact inference of (\ref{eq: bayesian_distribution}) is generally impractical, a particle filter is employed to estimate the hidden states $\mathbf{E_{t}}$, $\mathbf{I_{t}}$, $\mathbf{R_{t}}$ and $\mathbf{\Phi_{t}}$, given the observed variables $\mathbf{Z_{t}}$.

\subsection{Motion Prediction}

The computation of the predictions of the surrounding vehicles is inspired by the method proposed by \cite{althoffThesis}. The system dynamics are abstracted into Markov chains, where the state space $X$ and input space $U$ are discretized into intervals. The state space consists of longitudinal position $s$ and velocity $v$, and the input space represents the potential acceleration.

The transition probability matrices of the Markov chains for a time step $\Upsilon(\tau)$, and for a time interval $\Upsilon([0, \tau])$, where $\tau$ is the time increment, are computed offline with reachability analysis \cite{althoffThesis}, using the following differential equation as the vehicle's longitudinal dynamics:
\begin{align}
        &\dot{s} = v \nonumber\\
        &\dot{v} =  
        \begin{cases} 
            a^{max}u , & 0 < v < v^{max} \\
            0, & v \leq 0 \lor v \geq v^{max}
        \end{cases}
        \label{eq: motion_model}
\end{align}

\noindent  where  $a^{max}$ and $v^{max}$ are the maximum allowed acceleration and velocity, respectively, and $u$ is sampled for the discretized input space $U$.

The states probability distributions for future time steps $p(t_{k + 1})$ and time intervals $p(t_{k}, t_{k + 1})$ are computed as follows:

\begin{equation}
    \begin{gathered}
        p(t_{k+1}) = \Gamma(t_{k})\Upsilon(\tau)p(t_{k}) \\
        p(t_{k}, t_{k+1}) = \Upsilon([0,\tau])p(t_{k})
        \label{eq: markov_chains}
    \end{gathered}
\end{equation}

\noindent where $t_k$ is the discrete time step and $\Gamma(t_{k})$ is the input transition matrix that represents the transition probabilities between the input states. This matrix is computed for each prediction step, considering two parts: a matrix $\Psi$ that depicts the vehicle's inherent behavior when there are no priorities for certain input values (all inputs will eventually have the same probability) and a priority vector $\lambda$ that takes into account the dependencies between corridors, the road layout, and the velocity limitations. The dependencies between corridors are computed considering an interaction matrix that contains the probability of collision between two states given two acceleration inputs, one for each state. When this matrix is multiplied with the predictions of the corridor causing the dependency, it results in a probability vector that represents the percentage that an acceleration input is allowed at a given state. It is a costly operation since the interaction matrix is dense and its size depends on the number of states and inputs.

The predictions of each corridor are fused together with their probabilities, inferred with the DBN, to generate a motion grid. An example of the generated grid for the last instant of the time horizon is shown in Figure \ref{fig: motionGrid}.

\section{Learning model}
\label{sec: learning_models}


\begin{figure*}[!t]
    \centering
    \includegraphics[width=0.81\linewidth]{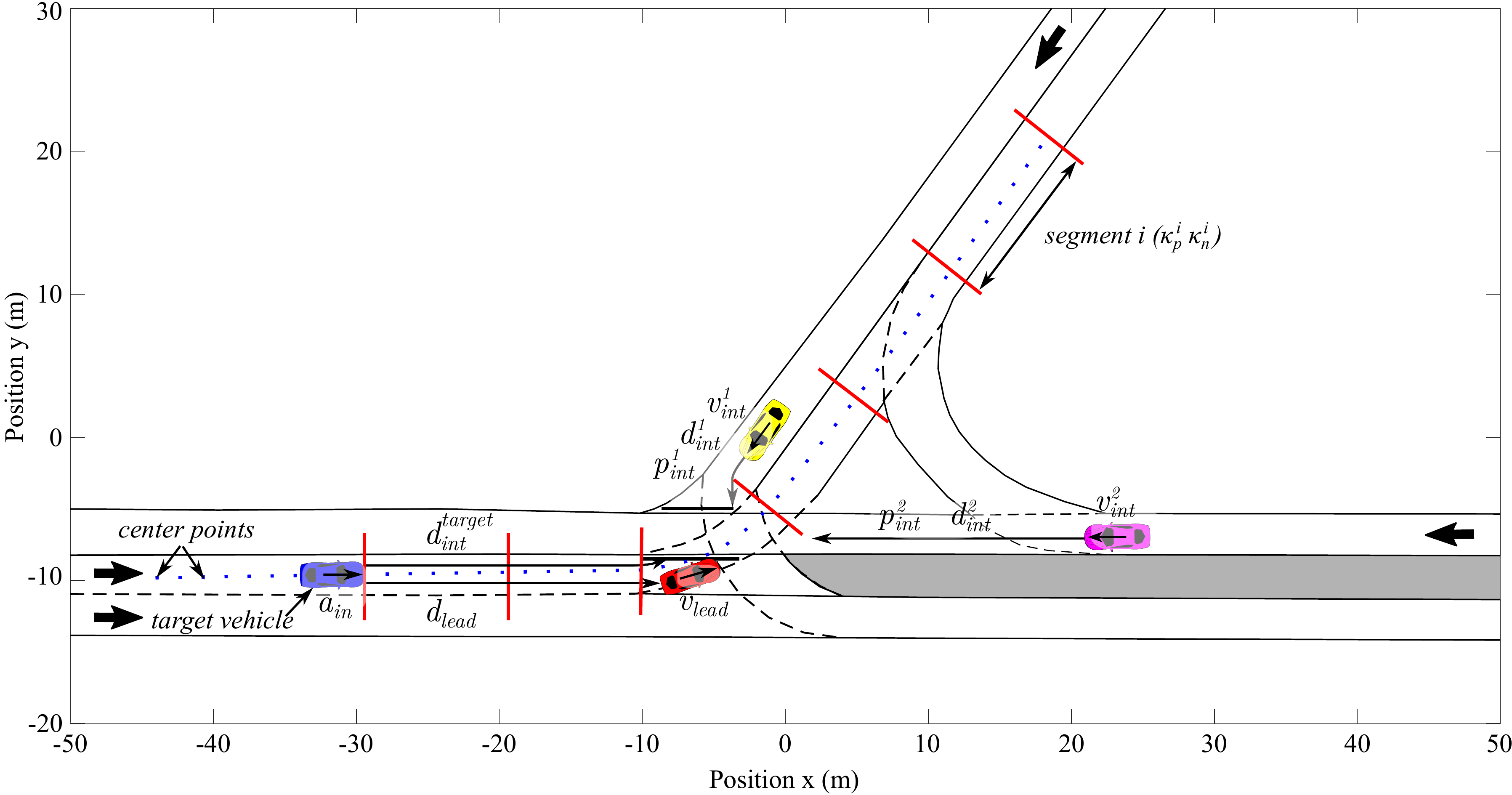}
    \caption{Input features example.}
    \label{fig: inputParam}
\end{figure*}

In this work, the aforementioned architecture is enhanced with the integration of a learning approach. This implementation tackles the limitation of the current motion prediction module, since it cannot adapt to different driving styles and driving scenarios. To do so, the priority vector $\lambda$ for the acceleration input is replaced by an acceleration profile generated as the output from a neural network model. These acceleration profiles take into account vehicle-to-vehicle and vehicle-to-layout interactions, as well as the history data from the target vehicle.

The replacement of the acceleration input greatly reduces the computational time of the system. Besides the time component, machine learning approaches can increase the accuracy of the predictions since the acceleration profile generated can self-adapt to different driving contexts and driver styles.

For the aforementioned task, an Auto-Regressive (AR) neural network \cite{Triebe} was selected due to its ability to describe time-varying processes and to predict future behavior based on past behavior. The AR model has been used in monitoring systems \cite{Yuting}, financial time series \cite{Ranaldi} and chemistry \cite{Barrett}.

The model takes as input the history of 4 s containing the acceleration $a_{in}$, the curvature of the path, the distance to the next intersection $d_{int}^{target}$, the distance to the leading vehicle $d_{lead}$ and its velocity $v_{lead}$, and the two vehicles that the most influence the path of the corridor in the next intersection (if exist), represented by their distance to the intersection $d_{int}^n$, their velocity $v_{int}^n$ and their priority $p_{int}^n$ with respect to the target vehicle. These inputs are represented in Figure \ref{fig: inputParam} and the variables are described in Table \ref{tab: input_features}. To obtain the curvature of the path, the centerline is equally divided into 6 segments and the positive and negative values of the curvature of each segment are integrated resulting in 12 discrete values ($\kappa_p^i$, $\kappa_n^i$), with $i \in \{1,\dotsc,6\}$ \cite{artunedo2020_machinelearningbased}.

To train the model, the Pytorch Lightning framework was used. For the optimization, the Adam algorithm \cite{adam} was selected, considering mean square error (MSE) and mean average error (MAE) as the losses. The learning rate was annealed in the form of a cosine curve every 10 epochs during the training procedure. 

The basic principle of the model is expressed in the following equation:

\begin{equation}
    y_t=Wx_{t-1}+ b + \epsilon_{t}
    \label{random progress}
\end{equation}

\noindent where $\epsilon_{t}$ is the white noise at  timestamp t, $x_{t-1}$ is the input state in timestamp $t-1$, $W$ and $b$ are the weight matrix and bias, respectively, that are needed to be updated. The output $y_t$ from the model is an acceleration time series representing 4 s into the future discretized at 0.1 s.

The datasets openDD \cite{openDD} and inD \cite{inD} were used to generate the input data to train the model. The whole dataset has been processed with the framework from Figure \ref{fig: flowchart} to generate the input for the training. 

\begin{table}[!t]
    \centering
    \caption{Input features variables.}
    \setlength\extrarowheight{6pt}
    \begin{tabular}{|M{3.5cm}|M{4cm}|} 
    \hline
      \textbf{Parameter}   &  \textbf{Variable}  \\ \hline
      \begin{tabular}{M{3.5cm}} target vehicle's acceleration \end{tabular} & $a_{in}$ \\ \hline
      \begin{tabular}{M{3.5cm}} distance to leading vehicle \end{tabular} & $d_{lead}$ \\ \hline
      \begin{tabular}{M{3.5cm}} leading vehicle's velocity \end{tabular} & $v_{lead}$ \\ \hline
      \begin{tabular}{M{3.5cm}} target vehicle's distance \\ to the intersection \end{tabular} & $d_{int}^{target}$ \\ \hline
      \begin{tabular}{M{3.5cm}} curvature parameters \end{tabular} & $(\kappa^i_p, \kappa^i_n), i \in \{1,\dotsc,6\}$  \\ \hline
      \begin{tabular}{M{3.5cm}} intersection parameters \end{tabular} & $(d_{int}^j, v_{int}^j, p_{int}^j), j \in \{1,2\}$ \\ \hline
      
    \end{tabular}
    
    \label{tab: input_features}
\end{table}
\renewcommand{\arraystretch}{1}

\section{Results}
\label{sec: experimental_results}


\subsection{Scenarios}
\label{sec: scenarios}

From the datasets inD \cite{inD} and INTERACTION \cite{interaction}, 4 situations were extracted following the procedures described in \cite{vinicius_access}. Table \ref{tab: scenarios} contains the information regarding the number of vehicles and duration of each situation followed by a brief description of each of them. 

\begin{table}[!t]
    \centering
    \caption{Information about the scenarios}
    \begin{tabular}{c|c|c}
         Situation & Number of vehicles & Duration \\ \hline
         A & 8 & 15.1 s  \\
         B & 6 & 19.8 s  \\
         C & 18 & 34.2 s \\
         D & 15 & 28.4 s
    \end{tabular}
    
    \label{tab: scenarios}
\end{table}

\begin{itemize}
    \item Situation A: four-armed intersection with two center left-turn lanes. The crossroad contains a lane with the right of way.
    \item Situation B: four-armed intersection where no lane has the right of way over the other.
    \item Situation C: T-junction intersection where the main road has the right of way and there is a left turn lane into the side road.
    \item Situation D: single-lane roundabout containing 3 entrances and 3 exits.

\end{itemize}

\subsection{Evaluation metrics}
\label{sec: metrics}

For the comparison with the physical-base model, two evaluation metrics have been implemented to evaluate the quality of the predictions: \textit{ADE} (Average Displacement Error) and \textit{FDE} (Final Displacement Error). Since the predictions are multi-modal, these metrics are used in the $min$ form, which means that for $k$ given corridors for the same vehicle, the minimum value is selected.

\begin{itemize}
    \item ADE: average $L_2$ distance between the ground truth positions and the weighted average position of the predictions.
    
    \begin{align*} 
        \text{ADE}_{k} &= \frac{1}{n}\sum_{i = 1}^{n}\sqrt{(x_{i}^{GT} - \hat{x}_{i})^{2} + (y_{i}^{GT} - \hat{y}_{i})^{2}} \\
        \text{mADE} &= \argmin_k \text{ADE}_{k} \\
    \end{align*}
    
    \item FDE: $L_2$ distance between the last ground truth position and the weighted average position of the last prediction.
    
    \begin{align*} 
        \text{FDE}_{k} &= \sqrt{(x_{n}^{GT} - \hat{x}_{n})^{2} + (y_{n}^{GT} - \hat{y}_{n})^{2}} \\
        \text{mFDE}  &= \argmin_k \text{FDE}_{k} \\
    \end{align*}
    
\end{itemize}

\noindent where $(x_{i}^{GT}, y_{i}^{GT})$ is the ground truth position obtained from the dataset,  $(\hat{x}_{i}, \hat{y}_{i})$ is the estimated position computed as the weighted average of the predictions and $n$ is the number of predictions.

\subsection{Comparison with baseline}

The model was evaluated in the situations described in Section \ref{sec: scenarios}. It was executed three times to take into account the stochastic nature of the framework, and the results here presented are the average of the executions.

\begin{figure}[!t]
    \centering
    \includegraphics[width=0.9\linewidth]{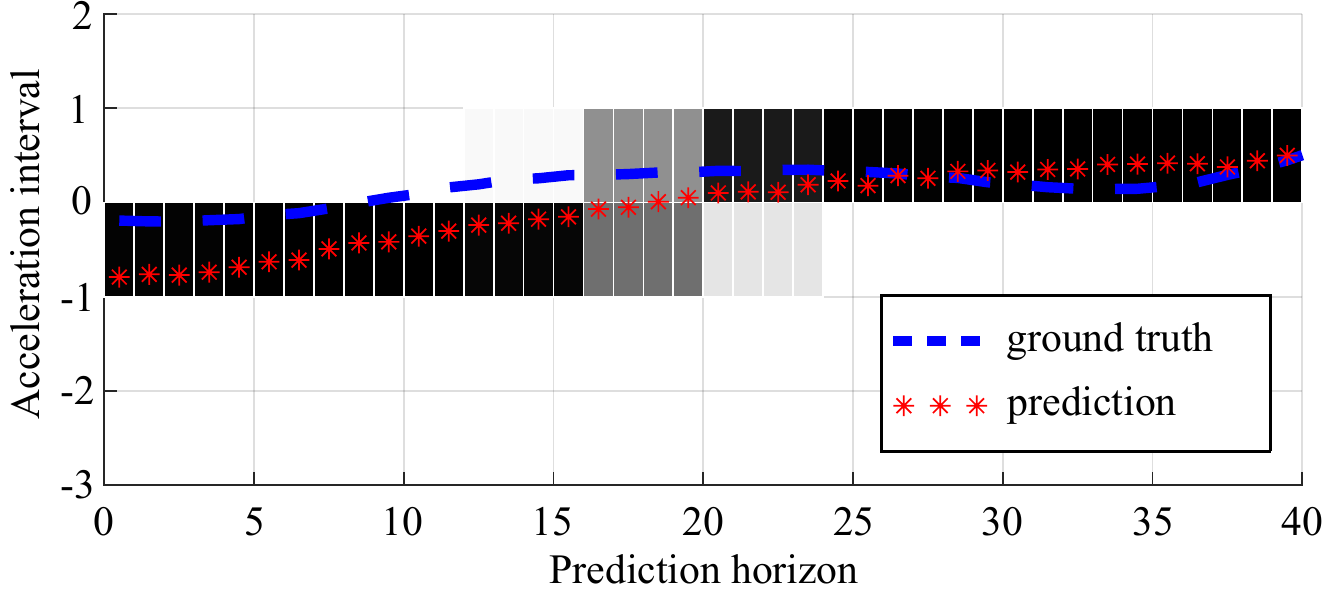} 
    \caption{Example of acceleration distribution obtained from the output of the model.}
    \label{fig: acc_distribution}
\end{figure}

In order to use the outputs from the model in the Markov chain-based motion prediction, the acceleration time series is converted into $n$ normal acceleration distributions. To represent the 4 s prediction horizon, 10 distributions are defined, each representing 0.4 s. These distributions are obtained by the evaluation of the input intervals (ranging from -3~m/s\textsuperscript{2} to 2~m/s\textsuperscript{2}) using a mean and a standard deviation obtained from the equally divided profile, as illustrated in Figure \ref{fig: acc_distribution}. In this figure, the blue line is the ground truth acceleration of the target vehicle, the red line represents the output $y_t$ from the AR model, and the black intensity corresponds to the probability of having this acceleration interval in a given prediction step inside the prediction horizon. Notice that this distribution is repeated in every state of the state space.

\begin{figure*}[!t]
    \centering
    \subfloat[]{\includegraphics[width=0.45\linewidth]{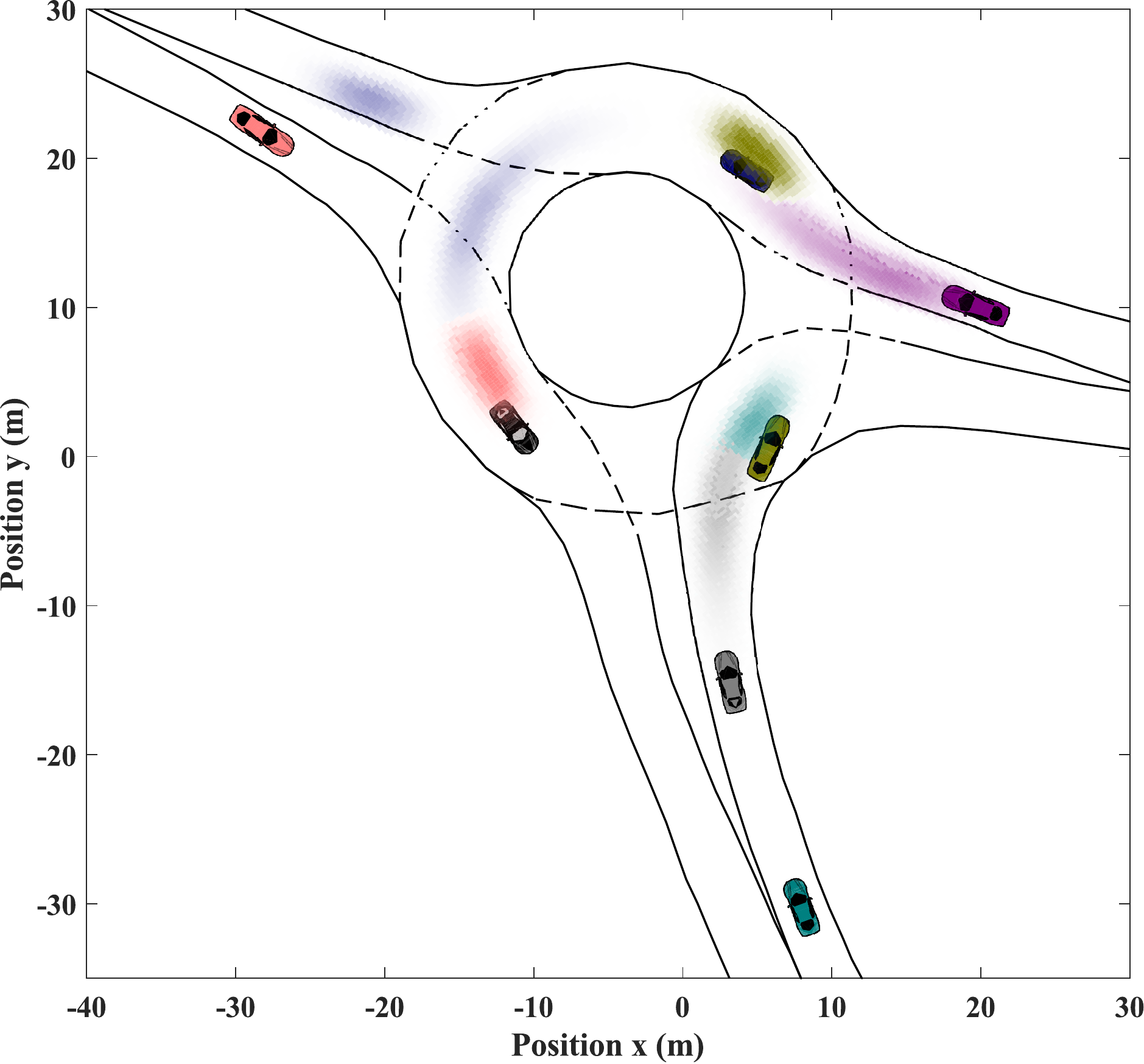}\label{fig: motionGrid}}
    \subfloat[]{\includegraphics[width=0.45\linewidth]{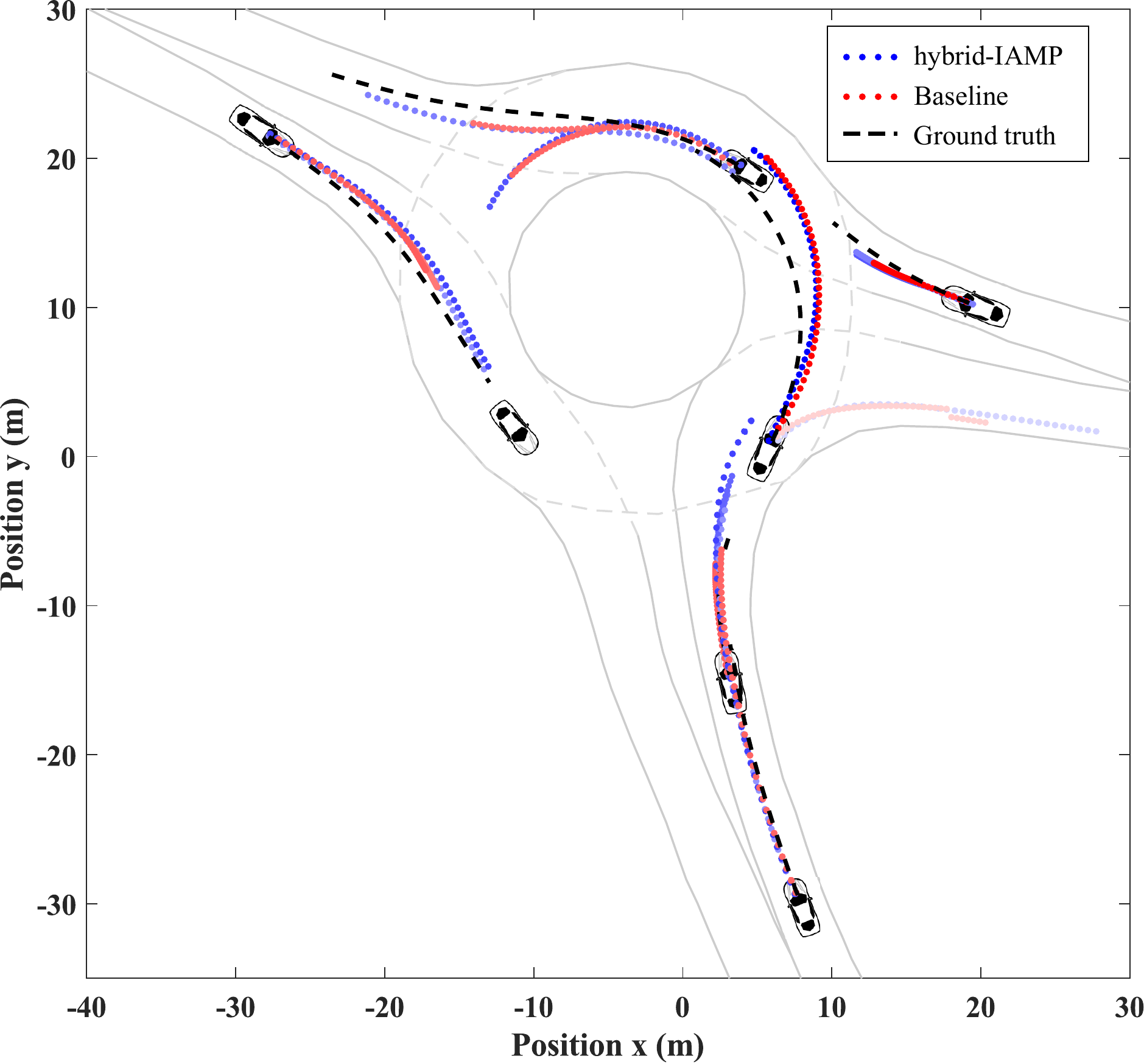}\label{fig: predictions_comp}}
    \caption{(a) Motion grid representing the predictions at 4.0 s in a given time step of the situation with hybrid-IAMP. (b) Comparison of the predictions: hybrid-IAMP (blue), baseline (red) and ground truth (black). The intensity of the predictions represents the probability of the corridor.}
\end{figure*}
 
\begin{table}[!t]
    \centering
    \caption{Comparison metrics for the instant from Figure \ref{fig: predictions_comp}.}
    \begin{tabular}{c|c|c}
         \backslashbox{Metric}{Model} & hybrid-IAMP & Baseline  \\ \hline
         mADE & 1.30 m & 2.10 m \\
         mFDE & 3.73 m & 5.74 m 
    \end{tabular}
    
    \label{tab: predMetrics}
\end{table}

\newcolumntype{g}{>{\columncolor{Gray}}c}
\begin{table}[!t]
    \centering
    \caption{Quantitative results.}
    \label{tab: results}
    \begin{tabular}{|c|c|c|c|c|}
    \hline 
    \backslashbox{Situation}{Model} & Hybrid-IAMP & Baseline \\ \hline
         A & \begin{tabular}{c|c}
                  time  & \textbf{0.97 s}  \\
                  mADE & \textbf{1.18 m}  \\
                  mFDE & \textbf{2.75 m}
             \end{tabular} & 
             \begin{tabular}{c|c}
                  time  & 3.83 s  \\
                  mADE & 1.43 m  \\
                  mFDE & 3.22 m
             \end{tabular}\\ \hline 

         B & \begin{tabular}{c|c}
                  time  & \textbf{1.14 s}  \\
                  mADE & \textbf{1.62 m}  \\
                  mFDE & \textbf{4.49 m}
             \end{tabular} & 
             \begin{tabular}{c|c}
                  time  & 5.96 s  \\
                  mADE & 1.92 m  \\
                  mFDE & 4.83 m
            \end{tabular}\\ \hline 

        C & \begin{tabular}{c|c}
                  time  & \textbf{0.80 s}  \\
                  mADE & \textbf{1.49 m}  \\
                  mFDE & 3.75 m
             \end{tabular} & 
             \begin{tabular}{c|c}
                  time  & 5.60 s  \\
                  mADE & 1.58 m  \\
                  mFDE & \textbf{3.73 m}
             \end{tabular}\\ \hline 
    
        D & \begin{tabular}{c|c}
                  time  & \textbf{1.22 s}  \\
                  mADE & \textbf{1.43 m}  \\
                  mFDE & \textbf{3.75 m}
             \end{tabular} & 
             \begin{tabular}{c|c}
                  time  & 3.18 s  \\
                  mADE & 1.96 m  \\
                  mFDE & 5.23 m
             \end{tabular}\\ \hline 
        \end{tabular}
\end{table}

An example of the predictions generated with the AR model (hybrid-IAMP) is presented in Figure \ref{fig: motionGrid}. In this figure, each footprint is the prediction for the last time step of the prediction horizon and corresponds to a vehicle with the same color. Notice that the prediction for the vehicle arriving from the right entrance respects the priority of the vehicle inside the roundabout and stays behind its prediction. The black vehicle is the ego vehicle. It also correctly predicts that the vehicle coming from the top left entrance can safely enter the roundabout.

The whole prediction horizon for this same instant is compared with the baseline and with the ground truth in Figure \ref{fig: predictions_comp}. As mentioned in Section \ref{sec: metrics}, the predictions in each time step are converted to the most probable position by computing a weighted average of the probabilistic footprint. Only vehicles having the ground truth position contained in the dataset are considered. As can be seen, the hybrid-IAMP can better adapt to different driving styles, and in general, generate a prediction closer to the ground truth when compared with the baseline. The metrics for this instant are presented in Table \ref{tab: predMetrics}.

Table \ref{tab: results} presents the prediction metrics for the hybrid-IAMP and for the baseline. These metrics are the average for the whole simulation time. It also includes the average time it took to compute the predictions. As can be seen, the hybrid implementation outperformed the baseline in both metrics, as well as in computational time.

\section{Concluding remarks}
\label{sec: conclusion}


In this paper, the framework IAMP was extended to include a learning mechanism able to generate personalized acceleration profiles that are used to compute the motion prediction of the surrounding vehicles. The hybrid-IAMP provided better performance in the predictions metrics and reduced computational cost when applied to four use cases extracted from the publicly available datasets \textit{inD} and \textit{INTERACTION}, proving that a hybrid approach can adapt better to different driving styles and driving scenarios.

As future work, more complex models will be implemented at this and other stages of the framework, always trying to improve the adaptability to the context while guaranteeing its interpretability.

\bibliographystyle{IEEEtran}
\bibliography{IEEEabrv,bibfile}

\vspace{12pt}

\end{document}